\documentclass[runningheads]{llncs}

\usepackage{graphicx}
\usepackage{comment}
\usepackage{amsmath,amssymb} 
\usepackage{color}
\usepackage{booktabs}
\usepackage{bm}
\usepackage{subcaption}
\newcommand{\eg}{\emph{e.g.}}
\newcommand{\ie}{\emph{i.e.}}
\newcommand{\etal}{\emph{et al.}}

\begin{document}
\pagestyle{headings}
\mainmatter
\def\ECCVSubNumber{6215}  

\title{Learning Video Representations by Transforming Time} 
\title{Learning to Represent Dynamics by Discriminating Time Distortions} 
\title{Learning Video Representations via\\ Discriminative Dynamic Time Warping}
\title{Learning Video Representations by\\ Discriminating Time Warps}
\title{Learning to Tell Motions Apart}
\title{Video Representation Learning by \\Learning to Tell Motions Apart}
\title{Video Representation Learning by \\ Recognizing Temporal Transformations}

%
\author{Simon Jenni\orcidID{0000-0002-9472-0425} \and 
Givi Meishvili\orcidID{0000-0002-0984-7078} \and \\
Paolo Favaro\orcidID{0000-0003-3546-8247}}

\authorrunning{S. Jenni et al.}
%

\institute{University of Bern, Switzerland\\
\email{\{simon.jenni,givi.meishvili,paolo.favaro\}@inf.unibe.ch}}

\maketitle

\begin{abstract}
We introduce a novel self-supervised learning approach to learn representations of videos that are responsive to changes in the motion dynamics. Our representations can be learned from data without human annotation and provide a substantial boost to the training of neural networks on small labeled data sets for tasks such as action recognition, which require to accurately distinguish the motion of objects.
We promote an accurate learning of motion without human annotation by training a neural network to discriminate a video sequence from its temporally transformed versions. To learn to distinguish non-trivial motions, the design of the transformations is based on two principles: 1) To define clusters of motions based on time warps of different magnitude; 2) To ensure that the discrimination is feasible only by observing and analyzing as many image frames as possible. Thus, we introduce the following transformations: forward-backward playback, random frame skipping, and uniform frame skipping.
Our experiments show that networks trained with the proposed method yield representations with improved transfer performance for action recognition on UCF101 and HMDB51.
\end{abstract}

\section{Introduction}

A fundamental goal in computer vision is to build representations of visual data that can be used towards tasks such as object classification, detection, segmentation, tracking, and action recognition  \cite{ImageNet,pascalch06,soomro2012ucf101,hmdb51}.
In the past decades, a lot of research has been focused on learning directly from single images and has done so with remarkable success \cite{redmon2016you,he2017mask,he2016deep}.
Single images carry crucial information about a scene. However, when we observe a temporal sequence of image frames, \ie, a video, it is possible to understand much more about the objects and the scene.
In fact, by moving, objects reveal their shape (through a change in the occlusions), their behavior (how they move due to the laws of Physics or their inner mechanisms), and their interaction with other objects (how they deform, break, clamp etc.). 
However, learning such information is non trivial. Even when labels related to motion categories are available (such as in action recognition), there is no guarantee that the trained model will learn the desired information, and will not instead simply focus on a single iconic frame and recognize a key pose or some notable features strongly correlated to the action \cite{schindler2008action}.

To build representations of videos that capture more than the information contained in a single frame, we pose the task of learning an accurate model of motion as that of learning to distinguish an unprocessed video from a temporally-transformed 
one. Since similar frames are present in both the unprocessed and transformed
sequence, the only piece of information that allows their discrimination is their temporal evolution. 
This idea has been exploited in the past 
\cite{fernando2017self,lee2017unsupervised,li2019joint,misra2016shuffle,wei2018learning}
and is also related to work in time-series analysis, where dynamic time warping is used as a distance for temporal sequences \cite{KenjiIwanaU2020}.

In this paper, we analyze different temporal transformations and evaluate how learning to distinguish them yields a representation that is useful to classify videos into meaningful action categories.
Our main finding is that the most effective temporal distortions are those that can be identified only by observing the largest number of frames. 
For instance, the case of substituting the second half of a video with its first half in reverse order, can be detected already by comparing just the 3 frames around the temporal symmetry. In contrast, distinguishing when a video is played backwards from when it is played forward \cite{wei2018learning} may require observing many frames.
Thus, one can achieve powerful video representations by using as pseudo-task the classification of temporal distortions that differ in their long-range motion dynamics.
\begin{figure}[t]
    \centering
    (a)
    \begin{subfigure}{0.9\textwidth}
        \centering
    \includegraphics[width=0.15\linewidth,trim=0 1cm 0 2cm, clip]{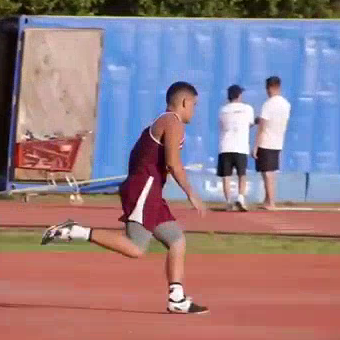}
    \includegraphics[width=0.15\linewidth,trim=0 1cm 0 2cm, clip]{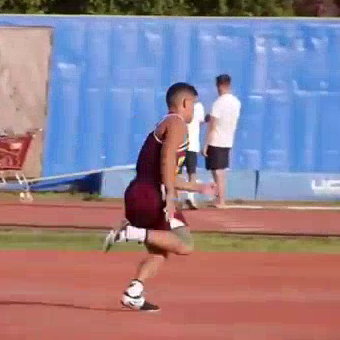}
    \includegraphics[width=0.15\linewidth,trim=0 1cm 0 2cm, clip]{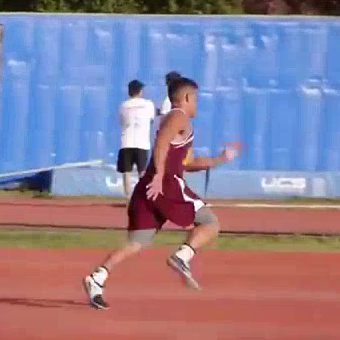}
    \includegraphics[width=0.15\linewidth,trim=0 1cm 0 2cm, clip]{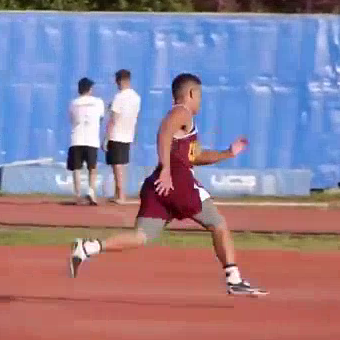}
    \includegraphics[width=0.15\linewidth,trim=0 1cm 0 2cm, clip]{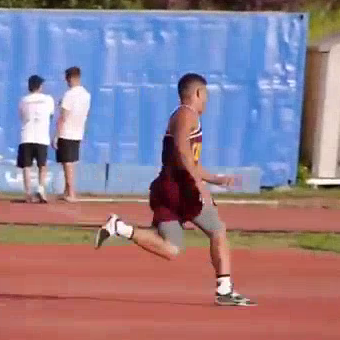}
    \includegraphics[width=0.15\linewidth,trim=0 1cm 0 2cm, clip]{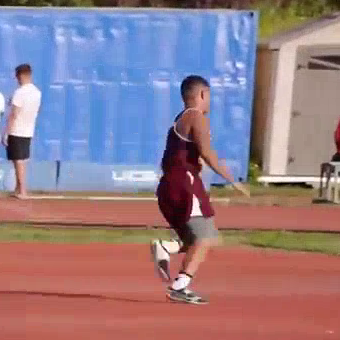}
    \vspace{-.3cm}
    \caption*{\tiny \phantom{0}0\hspace{0.14\linewidth}1\hspace{0.14\linewidth}\phantom{0}2\hspace{0.14\linewidth}\phantom{0}3\hspace{0.14\linewidth}\phantom{0}4\hspace{0.14\linewidth}\phantom{0}5}
    \vspace{.05cm}
    \end{subfigure}\\
    (b)
    \begin{subfigure}{0.9\textwidth}
        \centering
    \includegraphics[width=0.15\linewidth,trim=0 1cm 0 2cm, clip,trim=0 1cm 0 2cm, clip]{figures/sequence_sampling/0.png}
    \includegraphics[width=0.15\linewidth,trim=0 1cm 0 2cm, clip]{figures/sequence_sampling/2.png}
    \includegraphics[width=0.15\linewidth,trim=0 1cm 0 2cm, clip]{figures/sequence_sampling/4.png}
    \includegraphics[width=0.15\linewidth,trim=0 1cm 0 2cm, clip]{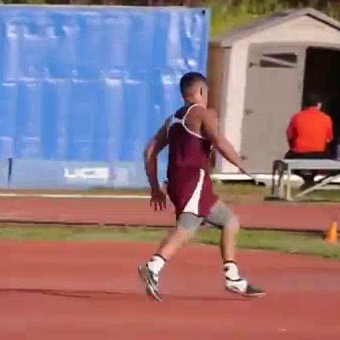}
    \includegraphics[width=0.15\linewidth,trim=0 1cm 0 2cm, clip]{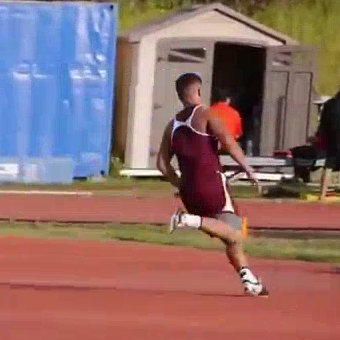}
    \includegraphics[width=0.15\linewidth,trim=0 1cm 0 2cm, clip]{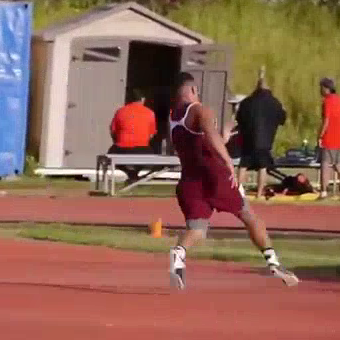}
    \vspace{-.3cm}
    \caption*{\tiny \phantom{0}0\hspace{0.14\linewidth}\phantom{0}2\hspace{0.14\linewidth}\phantom{0}4\hspace{0.14\linewidth}\phantom{0}6\hspace{0.14\linewidth}8\hspace{0.14\linewidth}10}
    \vspace{.05cm}
    \end{subfigure}\\
    (c)
    \begin{subfigure}{0.9\textwidth}
        \centering
    \includegraphics[width=0.15\linewidth,trim=0 1cm 0 2cm, clip]{figures/sequence_sampling/0.png}
    \includegraphics[width=0.15\linewidth,trim=0 1cm 0 2cm, clip]{figures/sequence_sampling/4.png}
    \includegraphics[width=0.15\linewidth,trim=0 1cm 0 2cm, clip]{figures/sequence_sampling/8.png}
    \includegraphics[width=0.15\linewidth,trim=0 1cm 0 2cm, clip]{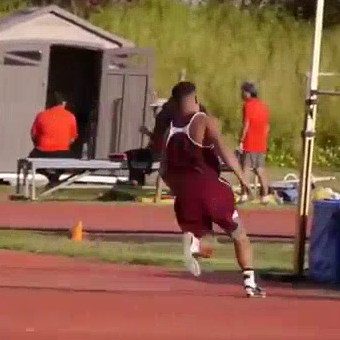}  
    \includegraphics[width=0.15\linewidth,trim=0 1cm 0 2cm, clip]{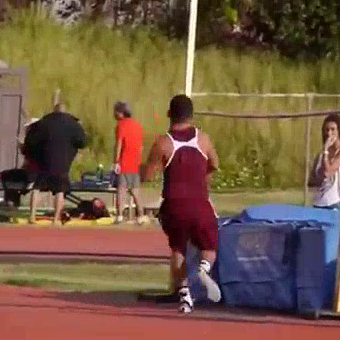}
    \includegraphics[width=0.15\linewidth,trim=0 1cm 0 2cm, clip]{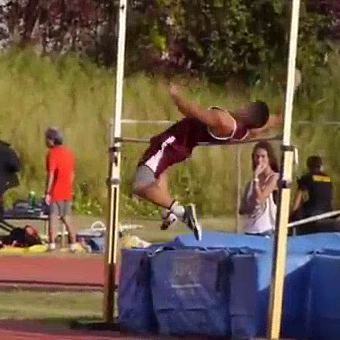}
    \vspace{-.3cm}
    \caption*{\tiny \phantom{0}0\hspace{0.14\linewidth}\phantom{0}4\hspace{0.14\linewidth}8\hspace{0.14\linewidth}12\hspace{0.14\linewidth}16\hspace{0.14\linewidth}20}
    \vspace{.05cm}
    \end{subfigure}\\
    (d)
    \begin{subfigure}{0.9\textwidth}
        \centering
    \includegraphics[width=0.15\linewidth,trim=0 1cm 0 2cm, clip]{figures/sequence_sampling/0.png}
    \includegraphics[width=0.15\linewidth,trim=0 1cm 0 2cm, clip]{figures/sequence_sampling/8.png}
    \includegraphics[width=0.15\linewidth,trim=0 1cm 0 2cm, clip]{figures/sequence_sampling/16.png}
    \includegraphics[width=0.15\linewidth,trim=0 1cm 0 2cm, clip]{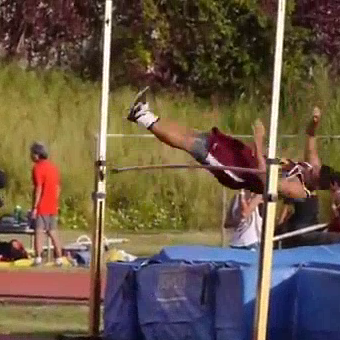}  \includegraphics[width=0.15\linewidth,trim=0 1cm 0 2cm, clip]{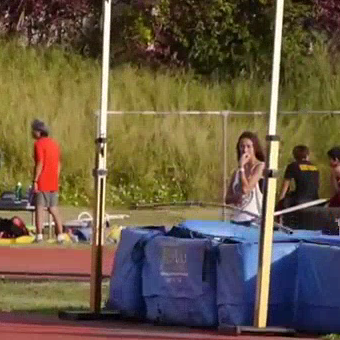}
    \includegraphics[width=0.15\linewidth,trim=0 1cm 0 2cm, clip]{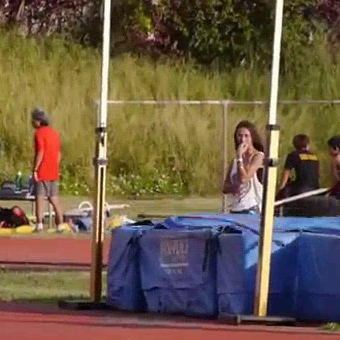}
    \vspace{-.3cm}
    \caption*{\tiny \phantom{0}0\hspace{0.14\linewidth}\phantom{0}8\hspace{0.14\linewidth}16\hspace{0.14\linewidth}24\hspace{0.14\linewidth}32\hspace{0.14\linewidth}40}
    \vspace{.05cm}
    \end{subfigure}\\
    (e)
    \begin{subfigure}{0.9\textwidth}
        \centering
    \includegraphics[width=0.15\linewidth,trim=0 1cm 0 2cm, clip]{figures/sequence_sampling/8.png}
    \includegraphics[width=0.15\linewidth,trim=0 1cm 0 2cm, clip]{figures/sequence_sampling/0.png}
    \includegraphics[width=0.15\linewidth,trim=0 1cm 0 2cm, clip]{figures/sequence_sampling/3.png}
    \includegraphics[width=0.15\linewidth,trim=0 1cm 0 2cm, clip]{figures/sequence_sampling/4.png}
    \includegraphics[width=0.15\linewidth,trim=0 1cm 0 2cm, clip]{figures/sequence_sampling/10.png}
    \includegraphics[width=0.15\linewidth,trim=0 1cm 0 2cm, clip]{figures/sequence_sampling/6.png}
    \vspace{-.3cm}
    \caption*{\tiny 8\hspace{0.14\linewidth}\phantom{0}0\hspace{0.14\linewidth}3\hspace{0.14\linewidth}\phantom{0}4\hspace{0.14\linewidth}10\hspace{0.14\linewidth}6}
    \vspace{.05cm}
    \end{subfigure}\\
    (f)
    \begin{subfigure}{0.9\textwidth}
        \centering
    \includegraphics[width=0.15\linewidth,trim=0 1cm 0 2cm, clip]{figures/sequence_sampling/0.png}
    \includegraphics[width=0.15\linewidth,trim=0 1cm 0 2cm, clip]{figures/sequence_sampling/8.png}
    \includegraphics[width=0.15\linewidth,trim=0 1cm 0 2cm, clip]{figures/sequence_sampling/16.png}
    \includegraphics[width=0.15\linewidth,trim=0 1cm 0 2cm, clip]{figures/sequence_sampling/24.png}
    \includegraphics[width=0.15\linewidth,trim=0 1cm 0 2cm, clip]{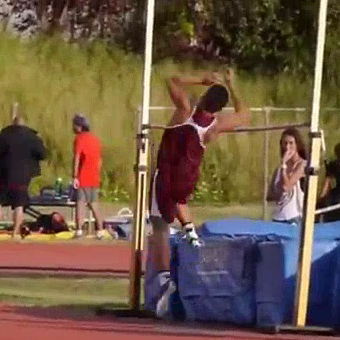}
    \includegraphics[width=0.15\linewidth,trim=0 1cm 0 2cm, clip]{figures/sequence_sampling/10.png}
    \vspace{-.3cm}
    \caption*{\tiny \phantom{0}0\hspace{0.14\linewidth}\phantom{0}8\hspace{0.14\linewidth}16\hspace{0.14\linewidth}24\hspace{0.14\linewidth}18\hspace{0.14\linewidth}\phantom{0}10}
    \vspace{.05cm}
    \end{subfigure}\\
    (g)
    \begin{subfigure}{0.9\textwidth}
        \centering
    \includegraphics[width=0.15\linewidth,trim=0 1cm 0 2cm, clip]{figures/sequence_sampling/0.png}
    \includegraphics[width=0.15\linewidth,trim=0 1cm 0 2cm, clip]{figures/sequence_sampling/5.png}
    \includegraphics[width=0.15\linewidth,trim=0 1cm 0 2cm, clip]{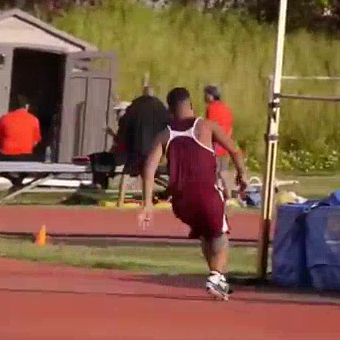}
    \includegraphics[width=0.15\linewidth,trim=0 1cm 0 2cm, clip]{figures/sequence_sampling/18.png}
    \includegraphics[width=0.15\linewidth,trim=0 1cm 0 2cm, clip]{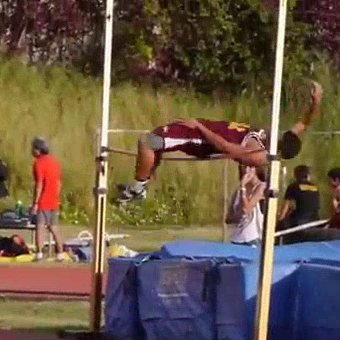}
    \includegraphics[width=0.15\linewidth,trim=0 1cm 0 2cm, clip]{figures/sequence_sampling/24.png}
    \vspace{-.3cm}
    \caption*{\tiny \phantom{0}0\hspace{0.14\linewidth}\phantom{0}5\hspace{0.14\linewidth}13\hspace{0.14\linewidth}18\hspace{0.14\linewidth}22\hspace{0.14\linewidth}24}
    \vspace{.05cm}
    \end{subfigure}
        \caption{\textbf{Learning from Temporal Transformations}. The frame number is indicated below each image. (a)-(d) \texttt{Speed} transformation by skipping: (a) 0 frames, (b) 1 frame, (c) 3 frames, and (d) 7 frames. (e) \texttt{Random}: frame permutation (no frame is skipped). (f) \texttt{Periodic}: forward-backward motion (at the selected speed). (g) \texttt{Warp}: variable frame skipping while preserving the order.}
    \label{fig:time-warps}
\end{figure}%
Towards this goal, we investigate 4 different temporal transformations of a video, which are illustrated in Fig.~\ref{fig:time-warps}:
\begin{enumerate}
    \item \textbf{Speed}: Select a subset of frames with uniform sub-sampling (\ie, with a fixed number of frames in between every pair of selected frames), while preserving the order in the original sequence;
    \item \textbf{Random}: Select a random permutation of the frame sequence;
    \item \textbf{Periodic}: Select a random subset of frames in their natural (forward) temporal order and then a random subset in the backward order;
    \item \textbf{Warp}: Select a subset of frames with a random sub-sampling (\ie, with a random number of frames in between every pair of selected frames), while preserving the natural (forward) order in the original sequence.
\end{enumerate}
We use these transformations to verify and illustrate the hypothesis that learning to distinguish them from one another (and the original sequence) is useful to build a representation of videos for action recognition.
For simplicity, we train a neural network that takes as input videos of the same duration and outputs two probabilities: One is about which one of the above temporal transformations the input sequence is likely to belong to
and the second is about identifying the correct speed of the chosen \textbf{speed} transformation.

In the Experiments section, we transfer features of standard 3D-CNN architectures (C3D \cite{tran2015learning}, 3D-ResNet \cite{hara2018can}, and R(2+1)D \cite{tran2018closer}) pre-trained through the above pseudo-task to standard action recognition data sets such as UCF101 and HMDB51, with improved performance compared to prior works.
We also show 
that features learned through our proposed pseudo-task capture long-range motion better than features obtained through supervised learning. 
Our project page \texttt{\url{https://sjenni.github.io/temporal-ssl}} provides code and additional experiments.

Our contributions can be summarized as follows:
1) We introduce a novel self-supervised learning task to learn video representations by distinguishing temporal transformations; 
2) We study the discrimination of the following novel temporal transformations: \textbf{speed}, 
    \textbf{periodic} and \textbf{warp};
3) We show that our features are a better representation of motion than features obtained through supervised learning and achieve state of the art transfer learning performance on action recognition benchmarks.

\section{Prior Work}

Because of the lack of manual annotation, our method belongs to self-supervised learning.
Self-supervised learning appeared in the machine learning literature more than 2 decades ago \cite{caruana1996promoting,ando2005predictive} and has been reformulated recently in the context of visual data with new insights that make it a promising method for representation learning \cite{Carl2015}.
This learning strategy is a recent variation on the unsupervised learning theme, which exploits labeling that comes for ``free'' with the data. Labels could be easily accessible and associated with a non-visual signal (for example, ego-motion \cite{agrawalCM15}, audio \cite{Owens_2018_ECCV}, text and so on), but also could be obtained from the structure of the data (\eg, the location of tiles \cite{Carl2015,noroozi2016unsupervised}, the color of an image \cite{zhang2016colorful,zhang2016split,larsson2017colorproxy}) or through transformations of the data \cite{gidaris2018unsupervised,jenni2018self,jenni2020steering}. 
Several works have adapted self-supervised feature learning methods from domains such as images or natural language to videos: Rotation prediction \cite{jing2018self}, Dense Predictive Coding \cite{han2019video}, and \cite{sun2019contrastive} adapt the BERT language model \cite{devlin2018bert} to sequences of frame feature vectors.

In the case of videos, we identify three groups of self-supervised approaches: 1) Methods that learn from videos to represent videos; 2) Methods that learn from videos to represent images; 3) Methods that learn from videos and auxiliary signals to represent both videos and the auxiliary signals (\eg, audio). 

\noindent\textbf{Temporal ordering methods.}
	Prior work has explored the temporal ordering of the video frames as a supervisory signal. For example, Misra \etal~\cite{misra2016shuffle} showed that learning to distinguish a real triplet of frames from a shuffled one yields a representation with temporally varying information (\eg, human pose). 
	This idea has been extended to longer sequences for posture and behavior analysis by using Long Short-Term Memories \cite{brattoli2017lstm}. 
	The above approaches classify the correctness of a temporal order directly from one sequence. 
	An alternative is to feed several sequences, some of which are modified, and ask the network to tell them apart \cite{fernando2017self}.
	Other recent work predicts the permutation of a sequence of frames \cite{lee2017unsupervised} or both the spatial and temporal ordering of frame patches  \cite{buchler2018improving,kim2019self}.
	Another recent work focuses on solely predicting the arrow of time in videos \cite{wei2018learning}. Three concurrent publications also exploit the playback speed as a self-supervision signal \cite{epstein2020oops,benaim2020speednet,yao2020video}.
	In contrast, our work studies a wider range of temporal transformations. Moreover, we show empirically that the temporal statistics extent  (in frames) captured by our features correlates to the transfer learning performance in action recognition.
	\\
\noindent\textbf{Methods based on visual tracking.}
The method of Wang and Gupta \cite{wang2015unsupervised} builds a metric to define similarity between patches. Three patches are used as input, where two patches are matched via tracking in a video and the third one is arbitrarily chosen. Tracking can also be directly solved during training, as shown in \cite{vondrick2018tracking}, where color is used as a supervisory signal. 
	By solving the task of coloring a grey-scale video (in a coherent manner across time), one can automatically learn how to track objects. 
Visual correspondences can also be learned by exploiting cycle-consistency in time \cite{wang2019learning} or by jointly performing region-level localization and fine-grained matching \cite{li2019joint}.
However, although trained on videos, these methods have not been used to build video representations or evaluated on action recognition.\\
\noindent\textbf{Methods based on auxiliary signals.}
Supervision can also come from additional signals recorded with images. For example, videos come also with audio.
The fact that the sounds are synchronized with the motion of objects in a video, already provides a weak supervision signal: One knows the set of possible sounds of visible objects, but not precisely their correspondence.
Owens \etal~\cite{owens2016ambient} show that, through the process of predicting a summary of ambient sound in video frames, a neural network learns a useful representation of objects and scenes.
Another way to learn a similar representation is via classification \cite{arandjelovic2017look}: A network is given an image-sound pair as input and must classify whether they match or not.
Korbar \etal~\cite{korbar2018cooperative} build audio and video representations by learning to synchronize audio and video signals using a contrastive loss.
Recently, \cite{patrick2020multi} use multi-modal data from videos also in a contrastive learning framework.
Several methods use optical flow as a supervision signal. 
For example, Wang \etal~\cite{wang2019self} extract motion and appearance statistics. Luo \etal~\cite{luo2017unsupervised} predict future atomic 3D flows given an input sequence, and Gan \etal~\cite{gan2018geometry} use geometry in the form of flow fields and disparity maps on synthetic and 3D movies. 
Optical flow is also used as input for video representation learning or filtering of the data \cite{wei2018learning}. 
Conversely, we do not make use of any auxiliary signals and learn video representations solely from the raw RGB frames.



\section{Learning Video Dynamics}

\begin{figure*}[t!]
    \centering
    \includegraphics[width=0.9\linewidth]{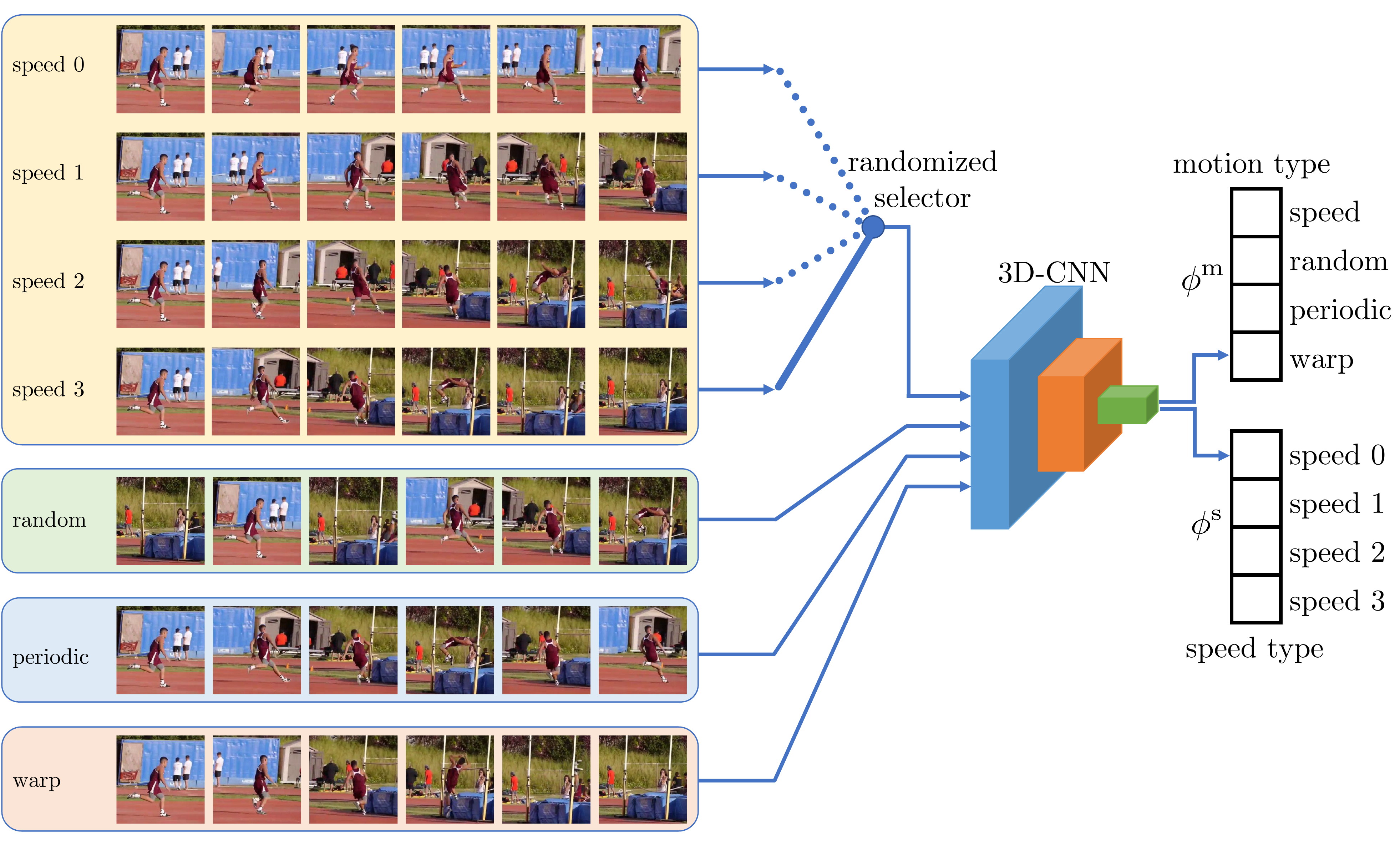}
    \caption{\textbf{Training a 3D-CNN to distinguish temporal transformations.} In each mini-batch we select a video speed (out of 4 possible choices), \ie, how many frames are skipped in the original video. Then, the 3D-CNN receives as input mini-batch a mixture of 4 possible transformed sequences: \texttt{speed} (with the chosen frame skipping), \texttt{random}, \texttt{periodic} and \texttt{warp}. The network outputs the probability of which motion type a sequence belongs to and the probability of which speed type the speed-transformed sequence has.}  
    \label{fig:transforms}
\end{figure*}

Recent work \cite{wang2019self} showed how a careful learning of motion statistics led to a video representation with excellent transfer performance on several tasks and data sets. The learning of motion statistics was made explicit by extracting optical flow between frame pairs, by computing flow changes, and then by identifying the region where a number of key attributes (\eg, maximum magnitude and orientation) of the time-averaged flow-change occurred.
In this work, we also aim to learn from motion statistics, but we focus entirely our attention on the temporal evolution without specifying motion attributes of interest or defining a task based on appearance statistics. We hypothesize that these important aspects could be implicitly learned and exploited by the neural network to solve the lone task of discriminating temporal transformations of a video. 
Our objective is to encourage the neural network to represent well motion statistics that require a long-range observation (in the temporal domain). 
To do so, we train the network to discriminate videos where the image content has been preserved, but not the temporal domain. For example, we ask the network to distinguish a video at the original frame rate from when it is played 4 times faster. 
Due to the laws of Physics, one can expect that, in general, \emph{executing} the same task at different speeds leads to different motion dynamics compared to when a video is just \emph{played} at different speeds (\eg, compare marching vs walking played at a higher speed). 
Capturing the subtleties of the dynamics of these motions requires more than estimating motion between 2 or 3 frames. Moreover, these subtleties are specific to the moving object, and thus they require object detection and recognition.

In our approach, we transform videos by sampling frames according to different schemes, which we call \emph{temporal transformations}.
To support our learning hypothesis, we analyze  transformations that require short- (\ie, temporally local) and long-range (\ie, temporally global) video understanding. As will be illustrated in the Experiments section, short-range transformations yield representations that transfer to action recognition with a lower performance than long-range ones.

\subsection{Transformations of Time}

Fig.~\ref{fig:transforms} illustrates how we train our neural network (a 3D-CNN \cite{tran2015learning}) to build a video representation (with 16 frames). 
In this section, we focus on the inputs to the network. As mentioned above, our approach is based on distinguishing different temporal transformations. We consider 4 fundamental types of transformations: Speed changes, random temporal permutations, periodic motions and temporal warp changes.
Each of these transformations boils down to picking a sequence of temporal indices to sample the videos in our data set.
${\cal V}_{\kappa}^\tau\subset \{0,1,2,\dots\}$ denotes the chosen subset of indices of a video based on the transformation $\tau\in\{0,1,2,3\}$ and with speed $\kappa$.\\
\noindent\textbf{Speed ($\tau=0$): }
In this first type we artificially change the video frame rate, \ie, its playing speed. We achieve that by skipping a different number of frames. 
We consider 4 cases, \textbf{Speed 0,~1,~2,~3} corresponding to $\kappa=0,~1,~2,~3$ respectively, where we skip $2^\kappa-1$ frames. The resulting playback speed of \textbf{Speed $\bm{\kappa}$} is therefore $2^\kappa$ times the original speed.
In the generation of samples for the training of the neural network we first uniformly sample $\kappa \in \{0,1,2,3\} $, the playback speed, and then use this parameter to define other transformations. This sequence is used in all experiments as one of the categories against either other speeds or against one of the other transformations below. The index sequence ${\cal V}^0_\kappa$ is thus $\rho+[0,1\cdot2^\kappa,2\cdot2^\kappa,\dots,15\cdot2^\kappa]$, where $\rho$ is a random initial index.\\ 
\noindent\textbf{Random ($\tau=1$): }
In this second temporal transformation we randomly permute the indices of a sequence without skipping frames. 
We fix $\kappa=0$ to ensure that the maximum frame skip between two consecutive frames is not too dissimilar to other transformations.   
This case is used as a reference, as random permutations can often be detected by observing only a few nearby frames. Indeed, in the Experiments section one can see that this transformation yields a low transfer performance. The index sequence ${\cal V}_0^\text{1}$ is thus $\rho+ \text{permutation}([0,1,2,\dots,15])$. This transformation is similar to that of the pseudo-task of Misra \etal\@ \cite{misra2016shuffle}. \\
\noindent\textbf{Periodic ($\tau=2$): }
This transformation synthesizes motions that exhibit approximate periodicity. To create such artificial cases we first pick a point $2\cdot2^\kappa<s<13\cdot2^\kappa$ where the playback direction switches. Then, we compose a sequence with the following index sequence: $0$ to $s$ and then from $s-1$ to $2s-15\cdot2^\kappa$. Finally, we sub-sample this sequence by skipping $2^\kappa-1$ frames. Notice that the randomization of the midpoint $s$ in the case of $\kappa>0$ yields pseudo-periodic sequences, where the frames in the second half of the generated sequence often do not match the frames in the first half of the sequence. The index sequence ${\cal V}^\text{2}_\kappa$ is thus $\rho+[0,1\cdot2^\kappa,2\cdot2^\kappa,\dots,\Bar{s}\cdot2^\kappa,(\Bar{s}-1)\cdot2^\kappa+\delta,\dots,(2\Bar{s}-15)\cdot2^\kappa+\delta])$, where $\Bar{s}=\lfloor s/2^\kappa \rfloor$, $\delta=s-\Bar{s}\cdot2^\kappa$, and $\rho=\max(0, (15-2\Bar{s})\cdot2^\kappa-\delta)$.\\
\noindent\textbf{Warp ($\tau=3$): }
In this transformation, we pick a set of $16$ ordered indices with a non-uniform number of skipped frames between them (we consider sampling any frame so we let $\kappa=0$). In other words, between any of the frames in the generated sequence we have a random number of skipped frames, each chosen independently from the set $\{0,\ldots,7\}$. This transformation creates a warping of the temporal dimension by varying the playback speed from frame to frame.
To construct the index sequence ${\cal V}^\text{3}_0$ we first sample the frame skips $s_j\in\{0,\ldots,7\}$ for $j=1,\ldots,15$ and set ${\cal V}^\text{3}_0$ to $\rho+[0,s_1, s_1+s_2, \dots,\sum_{j=1}^{15}s_j]$.


\subsection{Training}


Let $\phi$ denote our network, and let us denote with $\phi^\text{m}$ (\texttt{motion}) and $\phi^\text{s}$ (\texttt{speed}) its two softmax outputs (see Fig.~\ref{fig:transforms}). 
To train $\phi$ we optimize the following loss
\begin{align}
    -&\text{E}_{\kappa\sim{\cal U}[0,3],p\in{\cal V}^0_\kappa,q\in{\cal V}^1_0,s\in{\cal V}^2_\kappa,t\in{\cal V}^3_0, x}\Big[
    \log\big(\phi_0^\text{m}\left(x_p\right)\phi_1^\text{m}\left(x_q\right)\phi_2^\text{m}\left(x_s\right)\phi_3^\text{m}\left(x_t\right)\big)
    \Big]\\
        -&\text{E}_{\kappa\sim{\cal U}[0,3],p\in{\cal V}^0_\kappa,x}\big[ \log\left(\phi_\kappa^\text{s}\left(x_p\right)\right)\big]\nonumber
\end{align}
where $x$ is a video sample, the sub-index denotes the set of frames. This loss is the cross entropy both for \texttt{motion} and \texttt{speed} classification (see Fig.~\ref{fig:transforms}).

\subsection{Implementation}
Following prior work \cite{wang2019self}, we use the smaller variant of the C3D architecture \cite{tran2015learning} for the 3D-CNN transformation classifier in most of our experiments. 
Training was performed using the AdamW optimizer \cite{loshchilov2018decoupled} with parameters $\beta_1=0.9, \beta_2=0.99$ and a weight decay of $10^{-4}$. The initial learning rate was set to $3\cdot10^{-4}$ during pre-training and $5\cdot10^{-5}$ during transfer learning. The learning rate was decayed by a factor of $10^{-3}$ over the course of training using cosine annealing \cite{loshchilov2016sgdr} both during pre-training and transfer learning. We use batch-normalization \cite{ioffe2015batch} in all but the last layer. 
Mini-batches are constructed such that all the different coarse time warp types are included for each sampled training video. The batch size is set 28 examples (including all the transformed sequences).
The speed type is uniformly sampled from all the considered speed types. Since not all the videos allow a sampling of all speed types (due to their short video duration) we limit the speed type range to the maximal possible speed type in those examples. 
We use the standard pre-processing for the C3D network. In practice, video frames are first resized to $128\times 171$ pixels, from which we extract random crops of size $112 \times 112$ pixels. We also apply random horizontal flipping of the video frames during training. We use only the raw unfiltered RGB video frames as input to the motion classifier and do not make use of optical flow or other auxiliary signals.

\begin{table}[t]
\centering
\caption{\textbf{Ablation experiments.} We train a 3D-CNN to distinguish different sets of temporal transformations. The quality of the learned features is evaluated through transfer learning for action recognition on UCF101 (with frozen convolutional layers) and HMDB51 (with fine-tuning of the whole network). } \label{tab:ablation}
\resizebox{\textwidth}{!}{%
\begin{tabular}{@{}l@{\hspace{1em}}c@{\hspace{1em}}c@{\hspace{1em}}c@{}}
\toprule
 & \texttt{speed} & \textbf{UCF101} & \textbf{HMDB51} \\ 

\textbf{Pre-Training Signal} & loss & (conv frozen) & (conv fine-tuned) \\ \midrule

Action Labels UCF101           &    -     & 60.7\%      & 28.8\% \\ \midrule

Speed                               &    YES    & 49.3\%      & 32.5\% \\ \midrule
Speed + Random                      &    NO     & 44.5\%      & 31.7\% \\
Speed + Periodic                    &    NO     & 40.6\%      & 29.5\% \\
Speed + Warp                       &    NO     & 43.5\%      & 32.6\% \\ 
Speed + Random                      &    YES    & 55.1\%      & 33.2\% \\
Speed + Periodic                    &    YES    & 56.5\%      & 36.1\% \\
Speed + Warp                       &    YES    & 55.8\%      & 36.9\% \\ \midrule
Speed + Random + Periodic           &    NO     & 47.4\%      & 30.1\% \\
Speed + Random + Warp              &    NO     & 54.8\%      & 36.6\% \\
Speed + Periodic + Warp            &    NO     & 50.6\%      & 36.4\% \\ 
Speed + Random + Periodic           &    YES     & 60.0\%      & 37.1\% \\
Speed + Random + Warp              &    YES     & 60.4\%      & 39.2\% \\
Speed + Periodic + Warp            &    YES     & 59.5\%      & 39.0\% \\ \midrule
Speed + Random + Periodic + Warp   &    NO     & 54.2\%      & 34.9\% \\ 
Speed + Random + Periodic + Warp   &    YES    & 60.6\%      & 38.0\% \\ \bottomrule
\end{tabular}%
}
\end{table}

\section{Experiments}

\noindent\textbf{Datasets and Evaluation.} In our experiments we consider three datasets. Kinetics \cite{zisserman2017kinetics} is a large human action dataset consisting of around 500K videos. Video clips are collected from YouTube and span 600 human action classes. We use the training split for self-supervised pre-training. UCF101 \cite{soomro2012ucf101} contains around 13K video clips spanning 101 human action classes. HMDB51 \cite{hmdb51} contains around 5K videos belonging to 51 action classes. Both UCF101 and HMDB51 come with three pre-defined train and test splits. We report the average  performance over all splits for transfer learning experiments. We use UCF101 train split 1 for self-supervised pre-training. 
For transfer learning experiments we skip 3 frames corresponding to transformation \textbf{Speed 2}. For the evaluation of action recognition classifiers in transfer experiments we use as prediction the maximum class probability averaged over all center-cropped sub-sequences for each test video. More details are provided in the supplementary material.\\

\noindent\textbf{Understanding the Impact of the Temporal Transformations.}
We perform ablation experiments on UCF101 and HMDB51 where we vary the number of different temporal transformations the 3D-CNN is trained to distinguish. The 3D-CNN is pre-trained for 50 epochs on UCF101 with our self-supervised learning task. We then perform transfer learning for action recognition on UCF101 and HMDB51. On UCF101 we freeze the weights of the convolutional layers and train three randomly initialized fully-connected layers for action recognition. This experiment treats the transformation classifier as a fixed video feature extractor. On HMDB51 we fine-tune the whole network including convolutional layers on the target task. This experiment therefore measures the quality of the network initialization obtained through self-supervised pre-training. In both cases we again train for 50 epochs on the action recognition task. The results of the ablations are summarized in Table~\ref{tab:ablation}. For reference we also report the performance of network weights learned through supervised pre-training on UCF101.

We observe that when considering the impact of a single transformation across different cases, the types \textbf{Warp} and \textbf{Speed} achieve the best transfer performance. With the same analysis, the transformation \textbf{Random} leads to the worst transfer performance on average. We observe that \textbf{Random} is also the easiest transformation to detect (based on training performance -- not reported). As can be seen in Fig.~\ref{fig:time-warps} (e) this transformation can lead to drastic differences between consecutive frames. Such examples can therefore be easily detected by only comparing pairs of adjacent frames.
In contrast, the motion type \textbf{Warp} can not be distinguished based solely on two adjacent frames and requires modelling long range dynamics.  
We also observe that distinguishing a larger number of transformations generally leads to an increase in the transfer performance.  
The effect of the \textbf{speed} type classification is quite noticeable. It leads to a very significant transfer performance increase in all cases. This is also the most difficult pseudo task (based on the training performance -- not reported). Recognizing the speed of an action is indeed challenging, since different action classes naturally exhibit widely different motion speeds (\eg, ``applying make-up'' vs. ``biking''). This task might often require a deeper understanding of the physics and objects involved in the video. 
\begin{table}[t]
    \centering
    \caption{\textbf{Comparison to prior work on self-supervised video representation learning.} Whenever possible we compare to results reported with the same data modality we used, \ie, unprocessed RGB input frames. * are our reimplementations. }
    \label{tab:comparison}
    \resizebox{\linewidth}{!}{%
    \begin{tabular}{@{}l@{\hspace{1em}}c@{\hspace{1em}}c@{\hspace{1em}}c@{\hspace{1em}}c@{\hspace{1em}}c}
    \toprule
    \textbf{Method}      &  \textbf{Ref}  &  \textbf{Network} &   \textbf{Train Dataset}   &   \textbf{UCF101}    &   \textbf{HMDB51} \\ \midrule

    Shuffle\&Learn \cite{misra2016shuffle} & \cite{misra2016shuffle}  & AlexNet    &  UCF101    &   50.2\%  &  18.1\%    \\
    O3N \cite{fernando2017self} & \cite{fernando2017self}  & AlexNet    &  UCF101    &   60.3\%  &  32.5\%    \\
    AoT \cite{wei2018learning} & \cite{wei2018learning}  & VGG-16    &  UCF101    &   78.1\%  &  -    \\
    OPN \cite{lee2017unsupervised} & \cite{lee2017unsupervised}  & VGG-M-2048    &   UCF101   &  59.8\%   &  23.8\%  \\    
    DPC \cite{han2019video}   & \cite{han2019video} &  3D-ResNet34   &   Kinetics   &   75.7\%  & 35.7\%   \\
    SpeedNet \cite{benaim2020speednet}   & \cite{benaim2020speednet} &   S3D-G  &   Kinetics   &   81.1\%  & 48.8\%   \\
    AVTS \cite{korbar2018cooperative} (RGB+audio)  & \cite{korbar2018cooperative} &  MC3   &   Kinetics   &   85.8\%  & 56.9\%   \\
    \midrule
    Shuffle\&Learn \cite{misra2016shuffle}* & -  & C3D    &  UCF101    &   55.8\%  &  25.4\%    \\
    3D-RotNet \cite{jing2018self}*   & - &  C3D  &   UCF101   &   60.6\%  & 27.3\%    \\
    Clip Order \cite{xu2019self} & \cite{xu2019self}  & C3D    &  UCF101    &  65.6\%   &   28.4\%  \\
    Spatio-Temp \cite{wang2019self} & \cite{wang2019self}  & C3D    &   UCF101   &  58.8\%   &   32.6\%    \\    
    Spatio-Temp \cite{wang2019self} & \cite{wang2019self}  & C3D    &   Kinetics   &  61.2\%   &   33.4\%    \\  
    3D ST-puzzle \cite{kim2019self} & \cite{kim2019self}  & C3D    &  Kinetics    &  60.6\%   &   28.3\%  \\
    Ours & -  & C3D    &  UCF101    &  \underline{68.3\%}   &   \underline{38.4\%}  \\
    Ours & -  & C3D    &  Kinetics    &  \textbf{69.9\%}  &   \textbf{39.6\%}  \\
    \midrule
    3D ST-puzzle \cite{kim2019self} & \cite{kim2019self}  & 3D-ResNet18    &  Kinetics    &  65.8\%   &   33.7\%  \\
    3D RotNet \cite{jing2018self}   & \cite{jing2018self} &  3D-ResNet18  &   Kinetics   &   66.0\%  & 37.1\%    \\
    DPC \cite{han2019video}   & \cite{han2019video} &  3D-ResNet18   &   Kinetics   &   68.2\%  & 34.5\%   \\
    Ours   & - &  3D-ResNet18   &   UCF101   &   \underline{77.3\%}  & \underline{47.5\%}   \\
    Ours   & - &  3D-ResNet18   &   Kinetics   &   \textbf{79.3\%}  & \textbf{49.8\%}   \\
    \midrule  
    Clip Order \cite{xu2019self} & \cite{xu2019self}  &  R(2+1)D    &  UCF101    &  \underline{72.4\%}   &   30.9\%  \\
    PRP \cite{yao2020video} & \cite{yao2020video}  &  R(2+1)D    &  UCF101    &  72.1\%   &   \underline{35.0\%}  \\
    Ours   & - &  R(2+1)D   &   UCF101   &  \textbf{81.6\%}   & \textbf{46.4\%}   \\

    \bottomrule
    \end{tabular}}
\end{table}
Notice also that our pre-training strategy leads to a better transfer performance on HMDB51 than supervised pre-training using action labels. This suggests that the video dynamics learned through our pre-training generalize well to action recognition and that such dynamics are not well captured through the lone supervised action recognition. \\

\noindent\textbf{Transfer to UCF101 and HMDB51. }
We compare to prior work on self-supervised video representation learning in Table~\ref{tab:comparison}. A fair comparison to much of the prior work is difficult due to the use of very different network architectures and training as well as transfer settings. We opted to compare with some commonly used network architectures (\ie, C3D, 3D-ResNet, and R(2+1)D) and re-implemented two prior works \cite{misra2016shuffle} and \cite{jing2018self} using C3D. We performed self-supervised pre-training on UCF101 and Kinetics. C3D is pre-trained for 100 epochs on UCF101 and for 15 epoch on Kinetics. 3D-ResNet and R(2+1)D are both pre-trained for 200 epochs on UCF101 and for 15 epochs on Kinetics. We fine-tune all the layers for action recognition. Fine-tuning is performed for 75 epochs using C3D and for 150 epochs with the other architectures. 
When pre-training on UCF101 our features outperform prior work on the same network architectures. Pre-training on Kinetics leads to an improvement in transfer in all cases. \\
\begin{figure}[t]
    \centering
    (a)
    \begin{subfigure}{0.95\textwidth}
        \centering
    \includegraphics[width=0.15\linewidth,trim=0 1cm 0 2cm, clip]{figures/sequence_sampling/0.png}
    \includegraphics[width=0.15\linewidth,trim=0 1cm 0 2cm, clip]{figures/sequence_sampling/1.png}
    \includegraphics[width=0.15\linewidth,trim=0 1cm 0 2cm, clip]{figures/sequence_sampling/2.png}
    \includegraphics[width=0.15\linewidth,trim=0 1cm 0 2cm, clip]{figures/sequence_sampling/3.png}
    \includegraphics[width=0.15\linewidth,trim=0 1cm 0 2cm, clip]{figures/sequence_sampling/4.png}
    \includegraphics[width=0.15\linewidth,trim=0 1cm 0 2cm, clip]{figures/sequence_sampling/5.png}\\
    \includegraphics[width=0.15\linewidth,trim=0 1cm 0 2cm, clip]{figures/sequence_sampling/3.png}
    \includegraphics[width=0.15\linewidth,trim=0 1cm 0 2cm, clip]{figures/sequence_sampling/4.png}
    \includegraphics[width=0.15\linewidth,trim=0 1cm 0 2cm, clip]{figures/sequence_sampling/5.png}
    \includegraphics[width=0.15\linewidth,trim=0 1cm 0 2cm, clip]{figures/sequence_sampling/6.png}
    \includegraphics[width=0.15\linewidth,trim=0 1cm 0 2cm, clip]{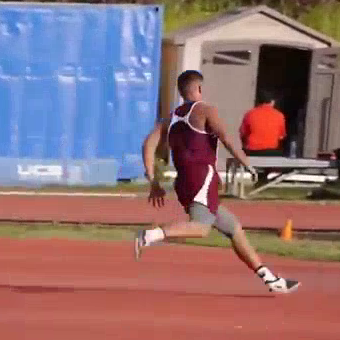}
    \includegraphics[width=0.15\linewidth,trim=0 1cm 0 2cm, clip]{figures/sequence_sampling/8.png}
    \end{subfigure}\\
    \vspace{.2cm}
    (b)
    \begin{subfigure}{0.95\textwidth}
        \centering
    \includegraphics[width=0.15\linewidth,trim=0 1cm 0 2cm, clip]{figures/sequence_sampling/0.png}
    \includegraphics[width=0.15\linewidth,trim=0 1cm 0 2cm, clip]{figures/sequence_sampling/1.png}
    \includegraphics[width=0.15\linewidth,trim=0 1cm 0 2cm, clip]{figures/sequence_sampling/2.png}
    \includegraphics[width=0.15\linewidth,trim=0 1cm 0 2cm, clip]{figures/sequence_sampling/3.png}
    \includegraphics[width=0.15\linewidth,trim=0 1cm 0 2cm, clip]{figures/sequence_sampling/4.png}
    \includegraphics[width=0.15\linewidth,trim=0 1cm 0 2cm, clip]{figures/sequence_sampling/5.png}\\
    \includegraphics[width=0.15\linewidth,trim=0 1cm 0 2cm, clip]{figures/sequence_sampling/7.png}
    \includegraphics[width=0.15\linewidth,trim=0 1cm 0 2cm, clip]{figures/sequence_sampling/8.png}
    \includegraphics[width=0.15\linewidth,trim=0 1cm 0 2cm, clip]{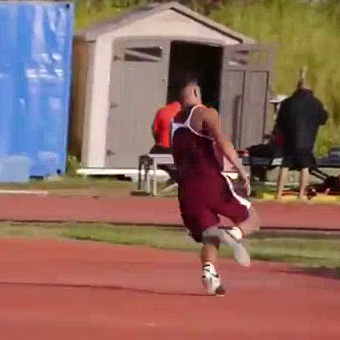}
    \includegraphics[width=0.15\linewidth,trim=0 1cm 0 2cm, clip]{figures/sequence_sampling/10.png}
    \includegraphics[width=0.15\linewidth,trim=0 1cm 0 2cm, clip]{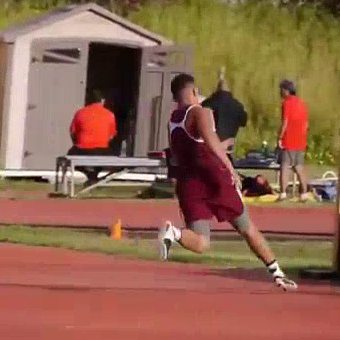}
    \includegraphics[width=0.15\linewidth,trim=0 1cm 0 2cm, clip]{figures/sequence_sampling/12.png}
    \end{subfigure}
        \caption{\textbf{Time-Related Pseudo-Tasks}. (a) Synchronization problem: The network is given two sequences with a time delay (4 frames in the example) and a classifier is trained to determine the delay. (b) The before-after problem: The network is given two non-overlapping sequences, and it needs to determine which comes first (the bottom sequence after the top one in the example).
        }
    \label{fig:timerelatedtasks}
\end{figure}%

\noindent\textbf{Long-Range vs Short-Range Temporal Statistics.}
To illustrate how well our video representations capture motion, we transfer them to other pseudo-tasks that focus on the temporal evolution of a video. One task is the classification of the synchronization of video pairs, \ie, how many frames one video is delayed with respect to the other. A second task is the classification of two videos into which one comes first temporally. These two tasks are illustrated in Fig.~\ref{fig:timerelatedtasks}.
In the same spirit, we also evaluate our features on other tasks and data sets and we report the results at our project page \texttt{\url{https://sjenni.github.io/temporal-ssl}}.

For the synchronization task, two temporally overlapping video sequences $x_1$ and $x_2$ are separately fed to the pre-trained C3D network to extract features $\psi(v_1)$ and $\psi(v_2)$ at the \texttt{conv5} layer. These features are then fused through $\psi(v_1)-\psi(v_2)$ and fed as input to a randomly initialized classifier consisting of three fully-connected layers trained to classify the offset between the two sequences. We consider random offsets between the two video sequences in the range -6 to +6.  
For the second task we construct a single input sequence by sampling two non-overlapping 8 frame sub-sequences $x_{i1}$ and $x_{i2}$, where $x_{i1}$ comes before $x_{i2}$. The network inputs are then either $(x_{i1}, x_{i2})$ for class ``before'' or $(x_{i2}, x_{i1})$ for the class ``after''. We reinitialize the fully-connected layers in this case as well. 
\begin{table}[t]
\centering
\caption{\textbf{Time-Related Pseudo-Tasks.} We examine how well features from different pre-training strategies can be transferred to time-related tasks on videos. As tasks we consider the synchronization of two overlapping videos and the temporal ordering of two non-overlapping videos. We report the accuracy on both tasks on the UCF101 test set and also report Mean Absolute Error (MAE) for the synchronization task. * are our reimplementations. \label{tab:timetask}}
\begin{tabular}{@{}l@{\hspace{2em}}c@{\hspace{1em}}c@{\hspace{2em}}c@{}}
\toprule
 & \multicolumn{2}{c}{\textbf{Sync.}} & \textbf{Before-After} \\ 
\textbf{Method}  &      Accuracy     &    MAE    &   Accuracy           \\ \midrule
Action Labels (UCF101)  &      36.7\%       &      \underline{1.85}       &     66.6\%           \\
3D-RotNet \cite{jing2018self}*  &      28.0\%       &      2.84       &     57.8\%           \\
Shuffle\&Learn \cite{misra2016shuffle}*  &     \underline{39.0\%}       &      1.89       &     \underline{69.8\%}           \\
Ours &     \textbf{42.4}\%        &      \textbf{1.61}       &            \textbf{76.9}\%                  \\ \bottomrule
\end{tabular}%
\end{table}
\begin{figure*}[!ht]
    \centering
    (a) 
    \begin{subfigure}{0.9\textwidth}
        \centering
        \includegraphics[width=0.95\linewidth]{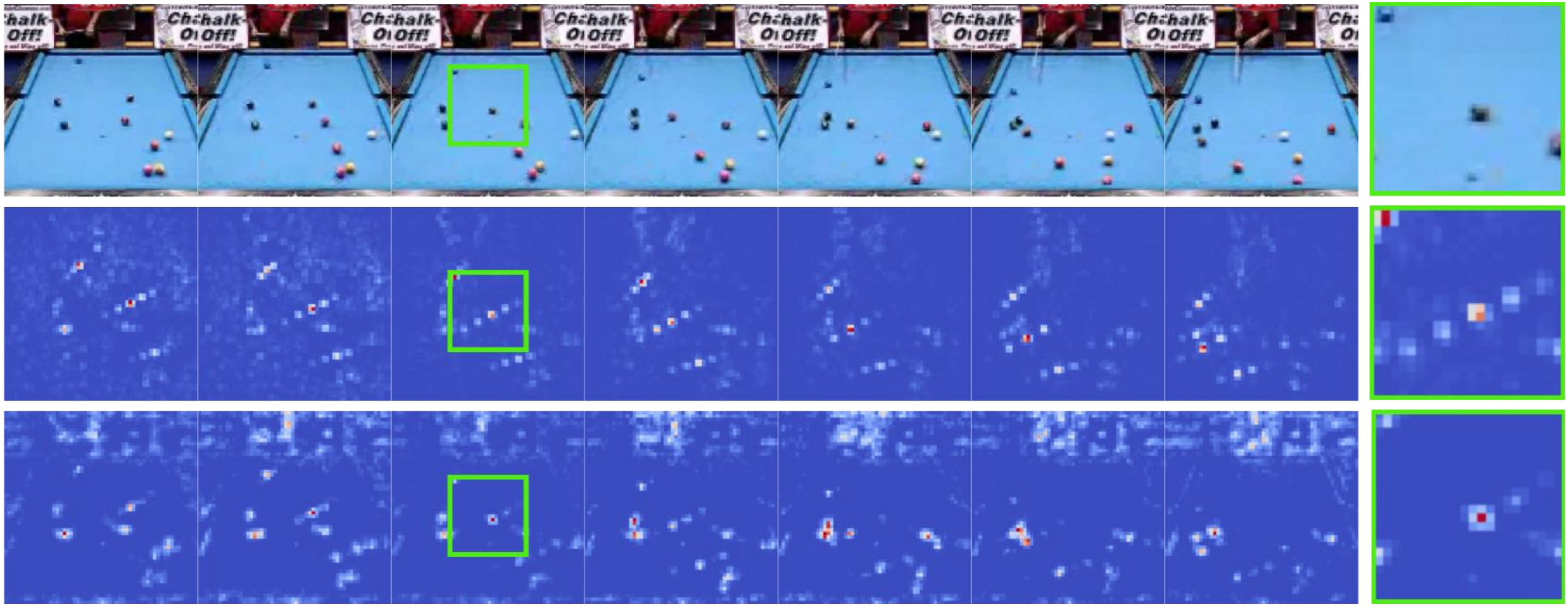}
    \end{subfigure}\\
    \vspace{.2cm}
    (b) 
    \begin{subfigure}{0.9\textwidth}
        \centering
        \includegraphics[width=0.95\linewidth]{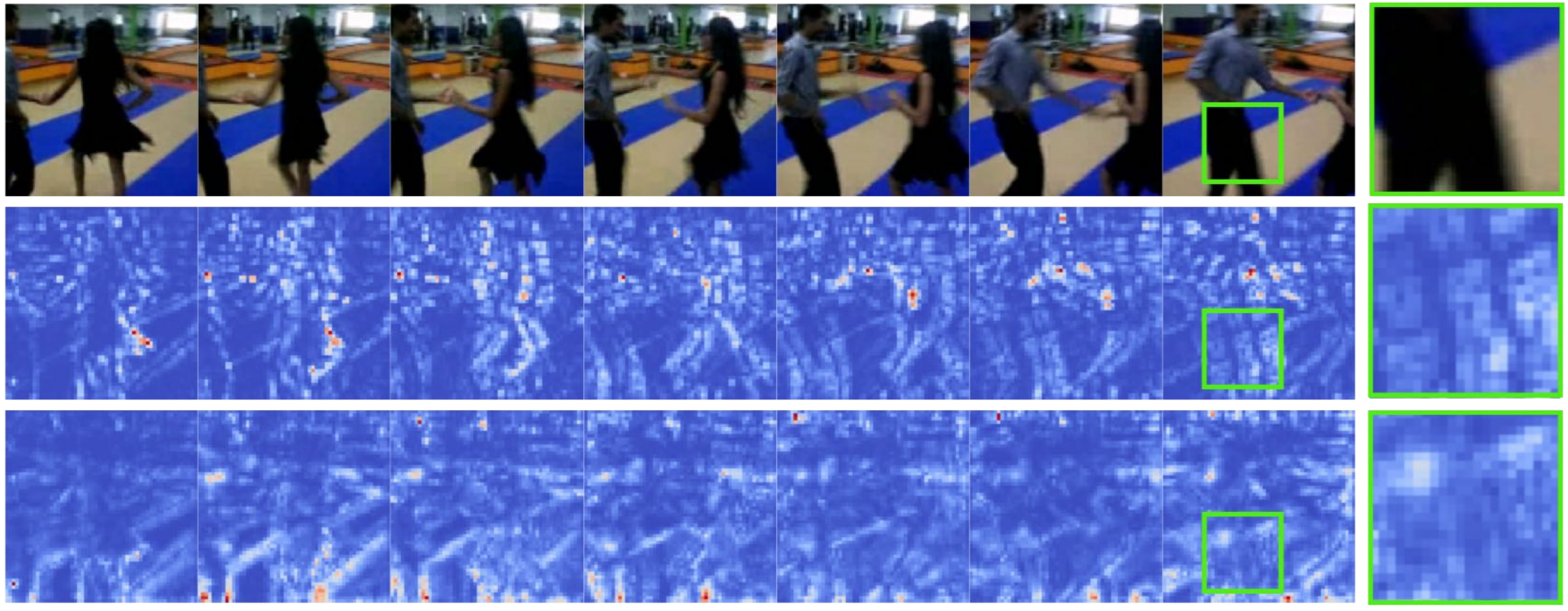}
    \end{subfigure}\\
    \vspace{.2cm}
    (c) 
    \begin{subfigure}{0.9\textwidth}
        \centering
        \includegraphics[width=0.95\linewidth]{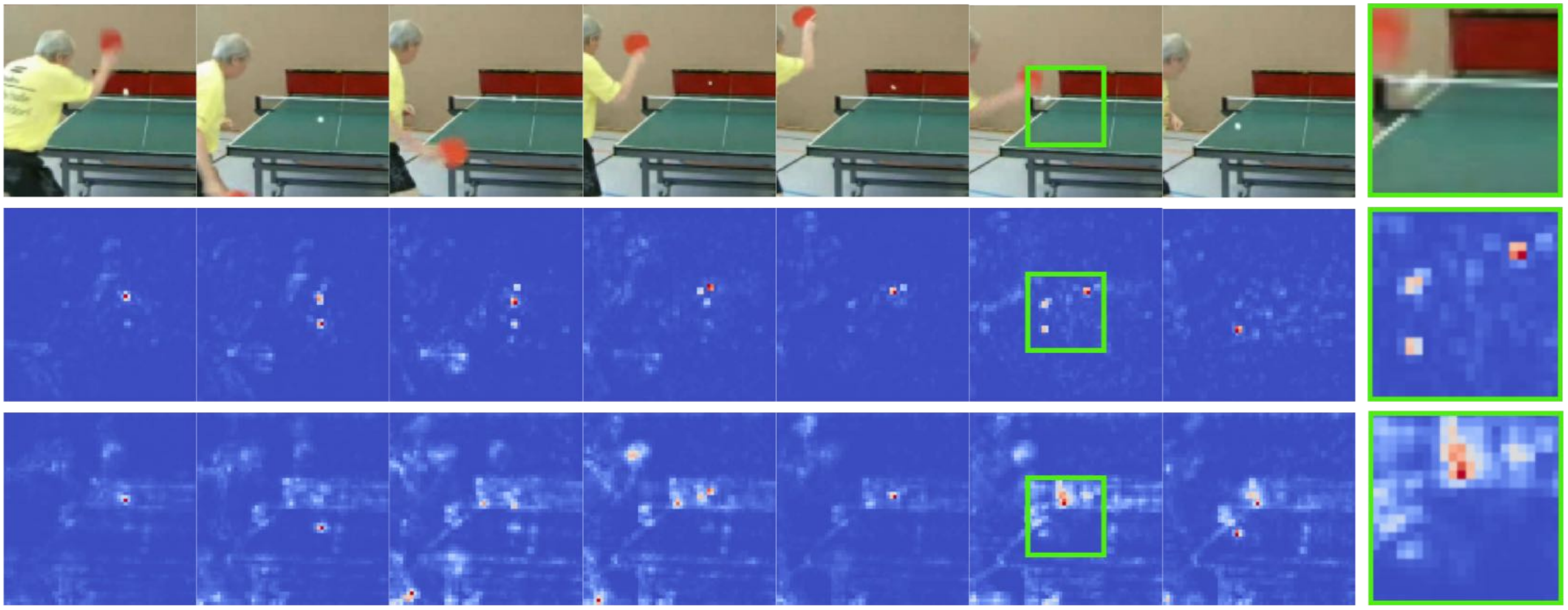}
    \end{subfigure}
    \caption{\textbf{Visualization of active pixels}. The first row in each block corresponds to the input video. Rows two and three show the output of our adaptation of Guided Backpropagation \cite{springenberg2014striving} when applied to a network trained through self-supervised learning and supervised learning respectively. In all three cases we observe that the self-supervised network focuses on image regions of moving objects or persons. In (a) we can also observe how long range dynamics are being detected by the self-supervised model. The supervised model on the other hand focuses a lot on static frame features  in the background. }
    \label{fig:grad_viz}
\end{figure*}

In Table~\ref{tab:timetask} we compare the performance of different pre-training strategies on the time-related pseudo-tasks. We see that our self-supervised features perform better at these tasks than supervised features and other self-supervised features, thus showing that they capture well the temporal dynamics in the videos.\\

\noindent\textbf{Visualization.} What are the attributes, factors or features of the videos that self-supervised and supervised models are extracting to perform the final classification? To examine what the self-supervised and supervised  models focus on, we apply Guided Backpropagation \cite{springenberg2014striving}. This method allows us to visualize which part of the input has the most impact on the final decision of the model. We slightly modify the procedure by subtracting the median values from every frame of the gradient video and by taking the absolute value of the result. We visualize the pre-trained self-supervised and supervised models on several test samples from UCF101. As one can see in Fig.~\ref{fig:grad_viz}, a model pre-trained on our self-supervised task tends to ignore the background and focuses on persons performing an action and on moving objects. Models trained with supervised learning on the other hand tend to focus more on the appearance of foreground and background. Another observation we make is that the self-supervised model identifies the location of moving objects/people in past and future frames. This is visible in row number 2 of blocks \textit{(a)} and \textit{(c)} of Fig.~\ref{fig:grad_viz}, where the network tracks the possible locations of the moving ping-pong and billiard balls respectively. 
A possible explanation for this observation is that our self-supervised task only encourages the learning of dynamics. The appearance of non-moving objects or static backgrounds are not useful to solve the pretext task and are thus ignored. \\

\noindent\textbf{Learning Dynamics vs. Frame Features.} The visualizations in Fig.~\ref{fig:grad_viz} indicate that features learned through motion discrimination focus on the dynamics in videos and not so much on static content present in single frames (\eg, background) when compared to supervised features. To further investigate how much the features learned through the two pre-training strategies rely on motion, we performed experiments where we remove all the dynamics from videos. To this end, we create input videos by replicating a single frame 16 times (resulting in a still video) and train the three fully-connected layers on \texttt{conv5} features for action classification on UCF101. Features obtained through supervised pre-training achieve an accuracy of 18.5\% (vs. 56.5\% with dynamics) and features from our self-supervised task achieve 1.0\% (vs. 58.1\%). Although the setup in this experiment is somewhat contrived (since the input domain is altered) it still illustrates that our features rely almost exclusively on motion instead of features present in single frames. This can be advantageous since motion features might generalize better to variations in the background appearance in many cases. \\ 

\noindent\textbf{Nearest-Neighbor Evaluation.}
We perform an additional quantitative evaluation of the learned video representations via the nearest-neighbor retrieval. The features are obtained by training a 3D-ResNet18 network on Kinetics with our pseudo-task and are chosen as the output of the global average pooling layer, which corresponds to a vector of size 512. For each video we extract and average features of 10 temporal crops. 
To perform the nearest-neighbor retrieval, we first normalize the features using the training set statistics. Cosine similarity is used as the metric to determine the nearest neighbors.
We follow the evaluation proposed by \cite{buchler2018improving} on UCF101. Query videos are taken from test split 1 and all the videos of train split 1 are considered as retrieval targets. A query is considered correctly classified if the $k$-nearest neighbors contain at least one video of the correct class (\ie, same class as the query). We report the mean accuracy for different values of $k$ and compare to prior work in Table~\ref{tab:nn}. Our features achieve state-of-the-art performance. \\
\begin{table}[t]
    \centering
    \caption{\textbf{Video Retrieval Performance on UCF101.} We compare to prior work in terms of $k$-nearest neighbor retrieval accuracy. Query videos are taken from test split 1 and retrievals are computed on train split 1. A query is correctly classified if the query class is present in the top-$k$ retrievals. We report mean retrieval accuracy for different values of $k$. }
    \label{tab:nn}
    \begin{tabular}{@{}l@{\hspace{1em}}c@{\hspace{1em}}c@{\hspace{1em}}c@{\hspace{1em}}c@{\hspace{1em}}c@{\hspace{1em}}c}
    \toprule
    \textbf{Method}      & \textbf{Network} &  \textbf{Top1}  &  \textbf{Top5} &   \textbf{Top10}   &   \textbf{Top20}    &   \textbf{Top50} \\ \midrule
    Jigsaw \cite{noroozi2016unsupervised} & AlexNet  &  19.7 & 28.5 & 33.5 & 40.0 & 49.4   \\ 
    OPN \cite{lee2017unsupervised} &  AlexNet  &   19.9 & 28.7 & 34.0 & 40.6 & 51.6 \\     
    B\"uchler \etal~\cite{buchler2018improving} &  AlexNet  &   \underline{25.7} & 36.2 & 42.2 &  49.2 & 59.5 \\     
    Clip Order \cite{xu2019self} &  R3D    &  14.1 & 30.3 & 40.0 & 51.1 & 66.5  \\
    SpeedNet \cite{benaim2020speednet}   &  S3D-G  &  13.0 & 28.1 & 37.5 & 49.5 & 65.0 \\
    PRP \cite{yao2020video} &  R3D    &  22.8  & \underline{38.5}  & \underline{46.7}  & \underline{55.2} & \underline{69.1} \\
    Ours &  3D-ResNet18  &  \textbf{26.1} & \textbf{48.5} & \textbf{59.1} & \textbf{69.6} & \textbf{82.8} \\     
    \bottomrule
    \end{tabular}
\end{table}

\begin{figure*}[t!]
    \centering
    \includegraphics[width=0.95\linewidth]{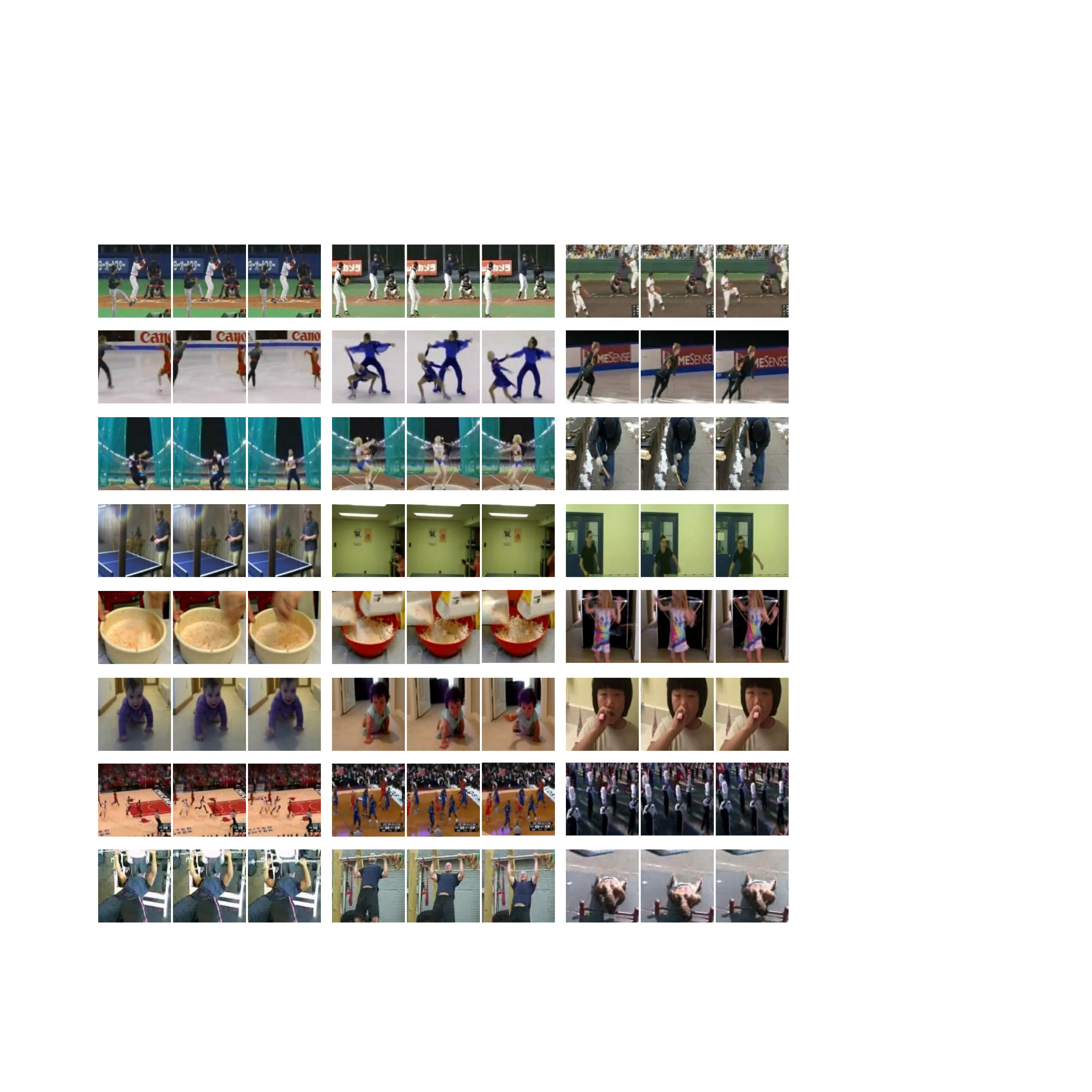}
    \caption{\textbf{Examples of Retrievals in UCF101.} Leftmost block of 3 columns: Frames from the query sequences. Second and third blocks of 3 columns: Frames from the two nearest neighbors.  }
    \label{fig:nn}
\end{figure*}

\noindent\textbf{Qualitative Nearest-Neighbor Results}
We show some examples of nearest neighbor retrievals in Fig.~\ref{fig:nn}. Frames from the query test video are shown in the leftmost block of three columns. The second and third blocks of three columns show the top two nearest neighbors from the training set. We observe that the retrieved examples often capture the semantics of the query well. This is the case even when the action classes do not agree (\eg, last row).

\section{Conclusions}
We have introduced a novel task for the self-supervised learning of video representations by distinguishing between different types of temporal transformations. This learning task is based on the principle that recognizing a transformation of time requires an accurate model of the underlying natural video dynamics. This idea is supported by experiments that demonstrate that features learned by distinguishing time transformations capture video dynamics more than supervised learning and that such features generalize well to classic vision tasks such as action recognition or time-related task such as video synchronization. \\ 
\noindent \textbf{Acknowledgements.} This work was supported by grants 169622\&165845 of the Swiss National Science Foundation.

\clearpage
\bibliographystyle{splncs04}
\bibliography{refs,egbib}
\end{document}